\documentclass[twocolumn,10pt]{article}

\usepackage[a4paper,margin=0.75in]{geometry}  
\usepackage{times}                            
\usepackage{titlesec}
 \usepackage{booktabs} 
\renewcommand{\thesection}{\arabic{section}}
\renewcommand{\thesubsection}{\thesection.\arabic{subsection}}
\renewcommand{\thesubsubsection}{\thesubsection.\arabic{subsubsection}}

\titleformat{\section}{\large\bfseries}{\thesection}{0.5em}{}
\titleformat{\subsection}{\normalsize\bfseries}{\thesubsection}{0.5em}{}
\titleformat{\subsubsection}{\normalsize\itshape}{\thesubsubsection}{0.5em}{}

\usepackage{graphicx}
\usepackage{amsmath, amssymb}
\usepackage{hyperref}
\usepackage{cite}

\renewcommand{\thesection}{\arabic{section}}
\renewcommand{\thesubsection}{\thesection.\arabic{subsection}}
\renewcommand{\thesubsubsection}{\thesubsection.\arabic{subsubsection}}

\makeatletter
\renewcommand{\p@section}{}
\renewcommand{\p@subsection}{\thesection.}
\usepackage{algorithm}
\usepackage{algorithmic}
\renewcommand{\p@subsubsection}{\thesubsection.}
\makeatother

\title{Cross Domain Evaluation of Multimodal Chain-of-Thought Reasoning of different datasets into the Amazon CoT Framework}

\author{
Nitya Tiwari\thanks{Equal contribution},
Parv Maheshwari\footnotemark[1],
Vidisha Agarwal\footnotemark[1] \\
Indian Institute of Technology Bombay, India \\
\texttt{\{22b2155, 22b1220, 22b1218\}@iitb.ac.in}
}
\date{}

\begin{document}

\maketitle

\begin{abstract}
 While recent work has extended CoT to multimodal settings, achieving state-of-the-art results on science question answering benchmarks like ScienceQA, the generalizability of these approaches across diverse domains remains underexplored. This work presents a comprehensive analysis of Multimodal Chain-of-Thought (Multimodal-CoT) reasoning, evaluating its effectiveness on the A-OKVQA, OKVQA and ChartQA datasets, which requires broad commonsense and world knowledge beyond scientific reasoning. We implement the two-stage framework proposed by Zhang et al.~\cite{zhang2023multimodal}, which separates rationale generation from answer inference and integrates vision features through a gated fusion mechanism with T5-based language models. Through systematic ablation studies, we analyze the contributions of vision features, rationale quality, and architectural choices. Our findings reveal that while vision integration significantly reduces hallucination in rationale generation, the effectiveness of CoT reasoning varies substantially across question types, with commonsense reasoning presenting particular challenges. This work provides practical insights for researchers implementing multimodal reasoning systems and identifies key areas for future improvement in cross-domain generalization.

\end{abstract}

\section{Introduction}

The remarkable success of large language models (LLMs) in recent years has been significantly enhanced by Chain-of-Thought (CoT) prompting~\cite{wei2022chain}, a technique that encourages models to generate intermediate reasoning steps before arriving at final answers. This approach has proven particularly effective for complex reasoning tasks, enabling models to decompose multi-step problems into more manageable sub-problems. However, most CoT research has focused exclusively on text-based reasoning, leaving the multimodal domain—where vision and language must be jointly processed—relatively unexplored.

Human reasoning naturally integrates information from multiple modalities. Consider reading a science textbook: our comprehension is greatly enhanced by the interplay between textual explanations and visual diagrams, figures, and tables. This multimodal nature of human knowledge acquisition motivates the development of AI systems capable of similar integrated reasoning. The challenge lies not only in processing multiple modalities but in generating coherent reasoning chains that effectively leverage both visual and textual information.

Recent work by Zhang et al.~\cite{zhang2023multimodal} introduced Multimodal-CoT, a framework that incorporates both language and vision modalities into a two-stage reasoning process. Their approach achieved state-of-the-art performance on the ScienceQA benchmark~\cite{lu2022learn}, demonstrating that models under 1 billion parameters can perform effective multimodal reasoning when properly structured. However, a critical question remains: \textit{How well does Multimodal-CoT generalize beyond scientific questions to broader domains requiring diverse forms of reasoning?}

\subsection{Motivation and Research Questions}

While the success of Multimodal-CoT on ScienceQA is impressive, scientific reasoning represents only one facet of human cognition. Real-world multimodal reasoning encompasses diverse challenges including commonsense understanding, world knowledge, spatial reasoning, and cultural context—domains where the characteristics of questions and required reasoning strategies may differ substantially from structured science problems.

This work addresses the following research questions:

\begin{enumerate}
    \item \textbf{Domain Generalization}: How does Multimodal-CoT perform on open-domain multimodal reasoning tasks that require commonsense and world knowledge rather than scientific principles?

    \item \textbf{Failure Modes}: What types of reasoning errors occur when applying Multimodal-CoT to diverse question types, and how do these differ from errors in scientific reasoning?
    
    \item \textbf{Practical Implementation}: What are the computational requirements and practical considerations for deploying Multimodal-CoT systems with limited resources?
\end{enumerate}


\section{Literature Review}
\label{sec:related}

\subsection{Chain-of-Thought Reasoning}

Chain-of-Thought (CoT) prompting emerged as a breakthrough technique for eliciting complex reasoning from large language models. Wei et al.~\cite{wei2022chain} demonstrated that providing few-shot examples with intermediate reasoning steps dramatically improves performance on arithmetic, commonsense, and symbolic reasoning tasks. This work established that sufficiently large models (typically exceeding 100 billion parameters) exhibit emergent reasoning abilities when prompted appropriately.

Subsequent research has explored various aspects of CoT reasoning:

\textbf{Zero-Shot CoT}: Kojima et al.~\cite{kojima2022large} showed that simply appending phrases like "Let's think step by step" to questions enables zero-shot CoT reasoning, eliminating the need for hand-crafted examples. This demonstrates that reasoning capabilities are latent within the models rather than purely learned from demonstrations.

\textbf{Self-Consistency}: Wang et al.~\cite{wang2022self} introduced self-consistency decoding, which samples multiple reasoning paths and selects the most consistent answer through majority voting. This approach significantly improves reliability by mitigating the impact of spurious reasoning chains.

\textbf{Least-to-Most Prompting}: Zhou et al.~\cite{zhou2022least} proposed decomposing complex problems into simpler sub-problems solved sequentially, with each solution building on previous results. This hierarchical approach proves particularly effective for compositional reasoning tasks.

\textbf{Automatic CoT Generation}: Zhang et al.~\cite{zhang2023automatic} developed Auto-CoT, which automatically constructs diverse demonstrations by clustering questions and generating reasoning chains using zero-shot CoT. This reduces the manual effort required for prompt engineering while maintaining effectiveness.

Despite these advances, CoT research has predominantly focused on text-only scenarios, with limited exploration of multimodal reasoning where visual information must be integrated with textual reasoning chains.

\subsection{Multimodal Learning and Vision-Language Models}

The integration of vision and language has been a longstanding challenge in artificial intelligence. Early approaches relied on separate vision and language modules with limited interaction~\cite{anderson2018bottom}. Recent years have witnessed the rise of unified vision-language models that process both modalities within a single framework.

\textbf{Vision-Language Pretraining}: Models like CLIP~\cite{radford2021learning}, ALIGN~\cite{jia2021scaling}, and BLIP~\cite{li2022blip} learn aligned vision-language representations through contrastive learning or image-text matching objectives. These models demonstrate strong zero-shot transfer capabilities across various vision-language tasks.

\textbf{Large Multimodal Models}: Recent work has scaled vision-language models to unprecedented sizes. Flamingo~\cite{alayrac2022flamingo} interleaves visual inputs with text in a few-shot learning framework. GPT-4V extends GPT-4's capabilities to visual inputs, while Gemini~\cite{reid2024gemini} provides native multimodal understanding. These models show impressive capabilities but require massive computational resources.

\textbf{Instruction-Tuned Multimodal Models}: LLaVA~\cite{liu2023visual}, InstructBLIP~\cite{dai2023instructblip}, and LLaMA-Adapter~\cite{zhang2023llama} fine-tune large vision-language models on instruction-following data, enabling more flexible and controllable multimodal reasoning. However, these approaches typically rely on models with billions of parameters.

A key limitation of current large multimodal models is their reliance on image captioning or visual feature extraction as an intermediate step, potentially losing fine-grained visual information critical for reasoning. Moreover, their computational requirements make deployment challenging for resource-constrained settings.

\subsection{Visual Question Answering}

Visual Question Answering (VQA) requires models to answer natural language questions about images. Traditional VQA approaches~\cite{antol2015vqa} focused on relatively simple questions answerable through object detection and attribute recognition.

\textbf{Attention Mechanisms}: Early neural VQA models employed attention mechanisms to align visual regions with question words. Methods like Bottom-Up Top-Down attention~\cite{anderson2018bottom}, Bilinear Attention Networks (BAN)~\cite{kim2018bilinear}, and Multi-modal Compact Bilinear pooling (MCB)~\cite{fukui2016multimodal} demonstrated the importance of fine-grained vision-language interaction.

\textbf{Knowledge-Based VQA}: Recognizing that many questions require external knowledge beyond what is visible in images, researchers developed knowledge-augmented VQA systems. OK-VQA~\cite{marino2019ok} and its successor A-OKVQA~\cite{schwenk2022okvqa} explicitly require commonsense and world knowledge, making them more challenging and realistic.

\textbf{Reasoning in VQA}: Recent work emphasizes multi-step reasoning for VQA. Visual Entailment~\cite{xie2019visual} frames VQA as a reasoning task. Compositional VQA datasets like CLEVR~\cite{johnson2017clevr} and GQA~\cite{hudson2019gqa} require structured reasoning over visual scenes. However, these approaches rarely generate explicit reasoning chains.

\subsection{Multimodal Chain-of-Thought}

The integration of CoT reasoning with multimodal inputs represents a nascent but promising research direction. Zhang et al.~\cite{zhang2023multimodal} introduced the first comprehensive study of CoT reasoning in multimodal settings, proposing a two-stage framework that separates rationale generation from answer inference.

Prior work on A-OKVQA has employed knowledge retrieval~\cite{gui2022kat}, caption-based reasoning with LLMs~\cite{lu2023chameleon}, and large multimodal models. However, none have applied the two-stage CoT framework with vision feature integration, making this a valuable testbed for evaluating Multimodal-CoT's generalization.

\subsection{Research Gap}

Our literature review reveals a significant research gap: while Multimodal-CoT has demonstrated strong performance on scientific reasoning, its applicability to open-domain multimodal reasoning remains unexplored. This work addresses this gap by:

\begin{enumerate}
    \item Implementing and evaluating Multimodal-CoT on A-OKVQA, OKVQA and ChartQA, datasets requiring diverse commonsense reasoning beyond scientific principles.

    \item Providing practical implementation guidance for researchers working with limited computational resources.
\end{enumerate}
\subsection{Motivation for Cross-Domain Evaluation}
Most prior work on Chain-of-Thought (CoT) prompting has focused on text-only reasoning
\cite{wei2022chain,kojima2022large,wang2022self}, while multimodal CoT remains relatively unexplored.
The Multimodal-CoT framework introduced by Zhang et al.~\cite{zhang2023multimodal} demonstrated that 
vision features fused with T5-based language models can achieve strong results on ScienceQA, a structured
scientific reasoning benchmark. However, ScienceQA represents only one domain—its questions have 
consistent structure, well-defined reasoning patterns, and multiple-choice answers.

To evaluate whether multimodal CoT generalizes to more diverse question distributions, we selected three
datasets that differ significantly in structure, modality complexity, and reasoning requirements:
ChartQA, OK-VQA, and A-OKVQA. Together, they allow us to probe numerical reasoning, commonsense
reasoning, and multimodal world-knowledge reasoning—domains that are radically different from ScienceQA.

\subsection{Why ChartQA?}
\textit{Numerical + Structured Visual Reasoning}\\
\begin{figure}
    \centering
    \includegraphics[width=1.0\linewidth]{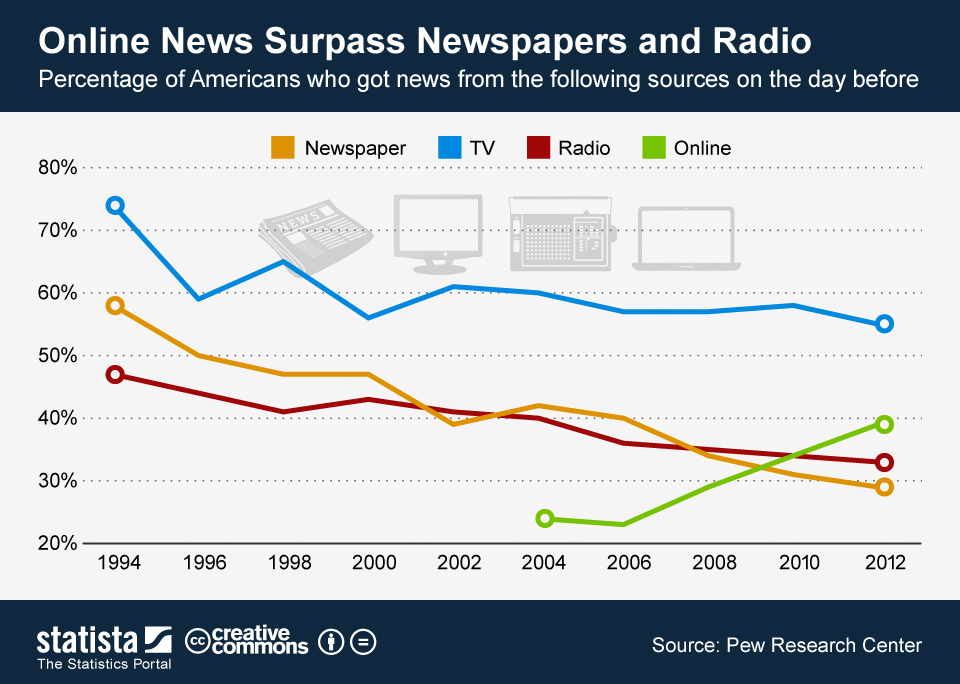}
    \caption{ChartQA Image}
    \label{fig:placeholder}
\end{figure}
\begin{figure}
    \centering
    \includegraphics[width=1.0\linewidth]{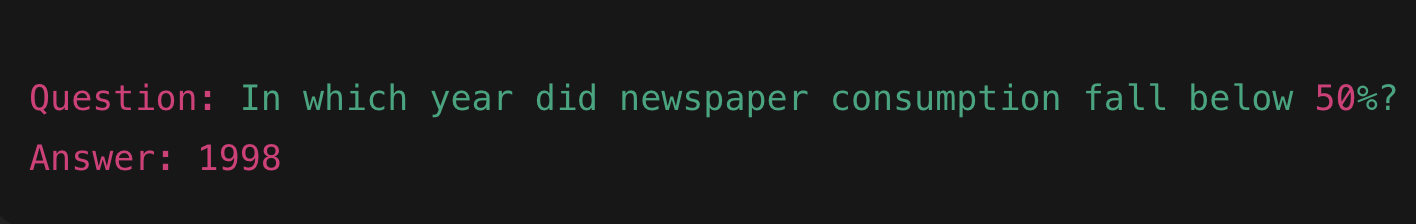}
    \caption{Chart QA Text}
    \label{fig:placeholder}
\end{figure}
ChartQA~\cite{chartqa} consists of bar charts, pie charts, and line plots paired with natural-language 
questions requiring numerical comparisons, trend understanding, and reasoning over visualized data.
Unlike natural images, charts encode information symbolically through axes, labels, and shapes.
This dataset tests whether Multimodal-CoT can:
\begin{itemize}
    \item interpret structured, synthetic visual layouts, \vspace{-8pt}
    \item perform arithmetic or ratio-based reasoning,  \vspace{-8pt}
    \item extract numeric answers from visual patterns rather than text.
\end{itemize}
Chart reasoning is fundamentally different from commonsense VQA, making it ideal for testing whether
CoT models can adapt to structured quantitative tasks.

\subsection{Why OK-VQA?}
\textit{Commonsense + External Knowledge Reasoning}\\ 

OK-VQA~\cite{marino2019ok} requires answering open-ended questions about natural images where the
answer is \emph{not} present in the image alone. Models must integrate:
\begin{itemize}
    \item commonsense knowledge,  \vspace{-8pt}
    \item factual/world knowledge,  \vspace{-8pt}
    \item indirect reasoning (e.g., “Why are people wearing coats?”).
\end{itemize}
Unlike ChartQA, OK-VQA answers are free-form, subjective, and knowledge-heavy. This tests whether 
Multimodal-CoT can generate rationales that incorporate both vision and background knowledge rather
than numerical inference.

\subsection{Why A-OKVQA?}
\textit{Ambiguous Open-Ended Reasoning + Rationales}\\

A-OKVQA~\cite{schwenk2022okvqa} extends OK-VQA by providing richer annotations and multiple 
valid rationales. It introduces:
\begin{itemize}
    \item multi-answer consensus scoring,  \vspace{-8pt}
    \item human-written explanations, \vspace{-8pt}
    \item improved question diversity,  \vspace{-8pt}
    \item tasks requiring multiple reasoning hops.
\end{itemize}
Because A-OKVQA includes rationales, it more directly aligns with CoT training and enables testing
whether multimodal rationales genuinely improve reasoning over ambiguous or knowledge-intensive images.
\begin{figure}
    \centering
    \includegraphics[width=0.8\linewidth]{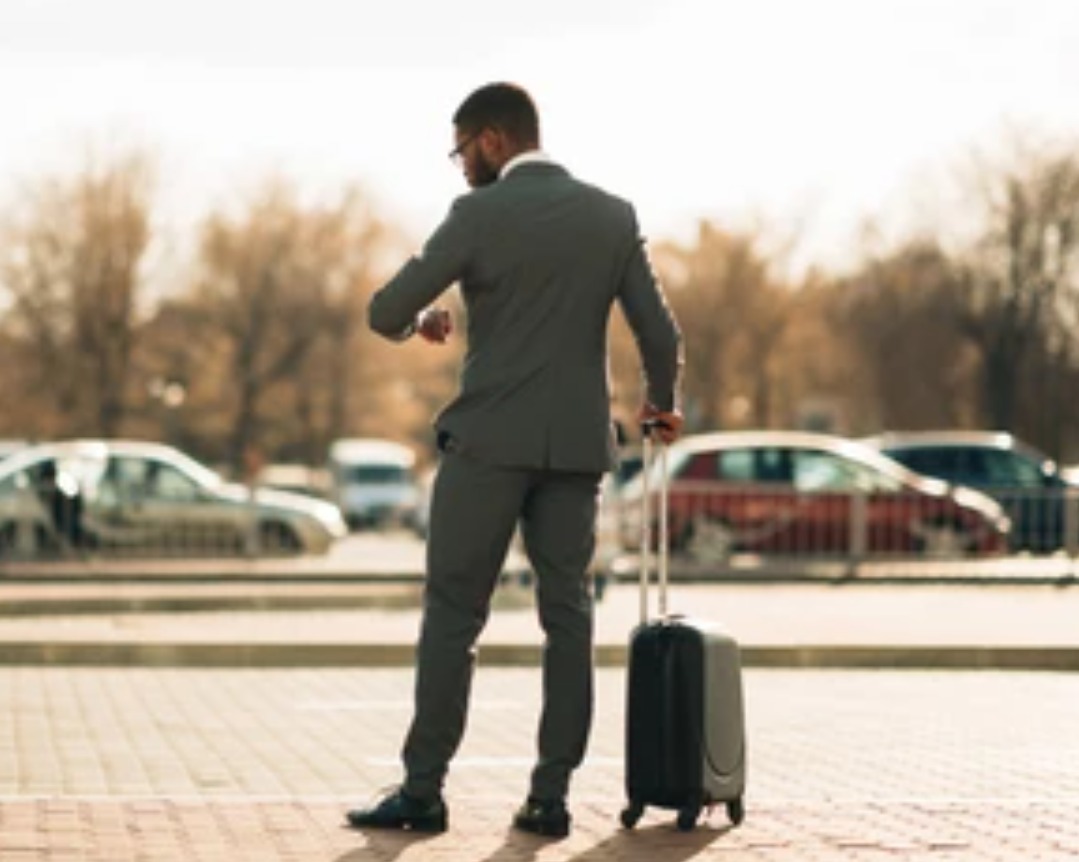}
    \caption{AOKVQA Image}
    \label{fig:placeholder}
\end{figure}
\begin{figure}
    \centering
    \includegraphics[width=0.9\linewidth]{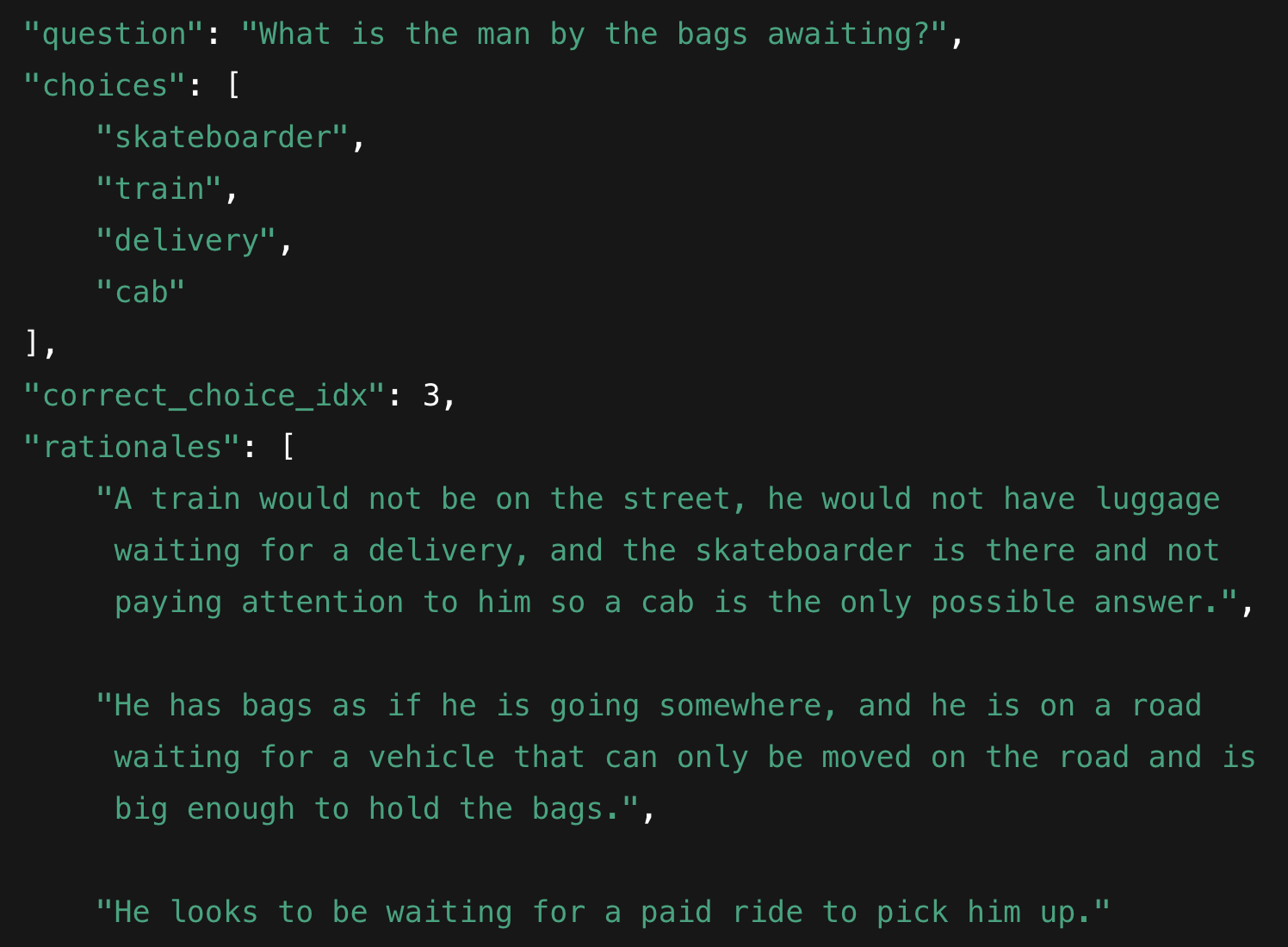}
    \caption{AOKVQA Text}
    \label{fig:placeholder}
\end{figure}
\begin{figure}
    \centering
    \includegraphics[width=0.8\linewidth]{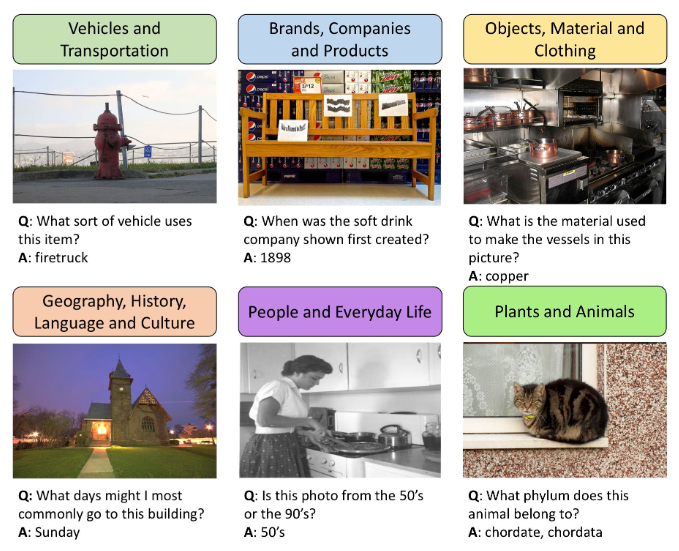}
    \caption{OKVQA Text}
    \label{fig:placeholder}
\end{figure}

\subsection{Comparison of the Three Datasets}
Individually, each dataset stresses a different aspect of multimodal reasoning:

\begin{itemize} \vspace{-5pt}
    \item \textbf{ChartQA:} Structured numeric reasoning from plots; requires recognizing bars, axes, 
    labels, and trends.  \vspace{-5pt}
    \item \textbf{OK-VQA:} Open-ended commonsense reasoning requiring external world knowledge.  \vspace{-5pt}
    \item \textbf{A-OKVQA:} Ambiguous, multi-annotator, rationale-rich reasoning combining vision, world knowledge, and natural language explanations. \vspace{-5pt}   
\end{itemize}
\vspace{5pt}
Together, they form a diverse testbed that moves well beyond ScienceQA’s structured format. Evaluating
MM-CoT on all three reveals whether the framework generalizes across numeric, commonsense, and 
knowledge-driven multimodal tasks.
\section{Methodology}
\subsection{OKVQA}
\subsubsection{Original Amazon CoT Pipeline}
The original Amazon CoT framework is designed for ScienceQA, leveraging multimodal data (images and text) in a multiple-choice format. The pipeline entails JSON-based structured data loaders, prompt constructors with predefined choices, answer extraction via regular expressions targeting single-letter responses, and evaluation using accuracy metrics. Model outputs are compared to ground-truth labels, with training checkpoints managed at epoch intervals.

\subsubsection{Framework Adaptations for OK-VQA}

We introduce the following key changes to enable open-ended visual question answering:

\begin{itemize}
    \item Implemented \texttt{load\_okvqa()} and \texttt{load\_okvqa\_img()} in \texttt{utils\_data.py} for parsing OK-VQA's question/annotation JSON.
    \item Extract majority-vote answers from multiple crowd annotations using normalized voting (Algorithm~\ref{alg:majvoting}).
    \item COCO image IDs are mapped to the existing vision feature pipeline.
    \item Train/val/test splits are maintained for compatibility.
\end{itemize}

\textbf{Algorithm: Majority Voting}
\begin{algorithm}
\label{alg:majvoting}
\begin{algorithmic}
\REQUIRE List of annotated answers $A = [a_1,...,a_m]$
\STATE Normalize $(a_i)$: lowercase + strip
\STATE Compute frequency of each normalized answer
\STATE Select highest-frequency answer as canonical
\STATE \textbf{return} canonical, original answers
\end{algorithmic}
\end{algorithm}

This gives us O($n$) data loading, robust answer aggregation, legacy vision feature compatibility.

\textbf{Dataset Class Extensions}
\textbf{Modifications:}
\begin{itemize}
    \item Extended \texttt{ScienceQADatasetStd} and \texttt{ScienceQADatasetImg} via inheritance to \texttt{OKVQADatasetStd} and \texttt{OKVQADatasetImg}.
    \item Integrated OK-VQA prompt formatting for open-answer questions.
\end{itemize}

Achieving Preservation of batching/tokenization, correct prompt templates, no overhead.

\subsubsection{Prompt Engineering for Open-Ended Answers}

\textbf{Modifications:}
\begin{itemize}
    \item Replaced single-letter answer regex with \texttt{extract\_open\_answer()} supporting multiple answer patterns.
    \item Normalizes output, removes trailing rationales.
    \item Modified prompt construction to allow empty choice lists for open-ended tasks.
\end{itemize}

\textbf{Mathematical Formulation:}
\begin{equation}
    a^* = \arg\max_{a \in A} \text{count}(\text{normalize}(a))
\end{equation}

This increases answer extraction robustness, supports variable model output formats.

\subsubsection{Evaluation Metrics Extension}

\textbf{Modifications:}
\begin{itemize}
    \item Exact Match (EM): String-normalized strict match.
    \item F1-score: Token-level overlap via precision/recall.
    \item Consensus scoring: Uses OK-VQA's fractional VQA-style voting.
\end{itemize}
Exact Match:
\begin{equation}
    \text{EM} = \mathbb{I}[normalize(pred) = normalize(gt)]
\end{equation}
F1-score:
\begin{equation}
    F_1 = \frac{2 \cdot Precision \cdot Recall}{Precision + Recall}
\end{equation}
Consensus Score:
\begin{equation}
 S = 
  \begin{cases}
    21.31 & \text{if matches} \geq 3\\
    0.6 & \text{if matches} = 2\\
    0.3 & \text{if matches} = 1\\
    0.0 & \text{otherwise}
  \end{cases} 
\end{equation}

Making the pipeline more appropriate for open-ended tasks, improving accuracy reporting.

\subsubsection{Training Infrastructure Enhancements}

Implemented step-based checkpointing (every 500 steps). Best 3 model checkpoints retained. Enabled resuming with \texttt{--resume\_from\_checkpoint}. This makes it resilient to interruptions, efficient model retention.

\begin{itemize}\setlength{\itemsep}{0.01em}
    \item Warmup period: 10\% of steps, minimum 100.
    \item Cosine decay after warmup.
\end{itemize}
\textbf{Equation:}
$$
LR(t) =
    \begin{cases}
        LR_{max} \cdot \frac{t}{W}, & t \leq W\\
        LR_{max} \cdot 0.5 \left[ 1 + \cos \left( \frac{\pi (t-W) }{ T-W } \right) \right], & t > W
    \end{cases}
$$
where $W$ is warmup steps, $T$ is total steps. Supports arbitrary effective batch size via \texttt{--gradient\_accumulation\_steps}.

\textbf{Equation:}
$$
\text{Effective Batch Size} = B_p \times N_{gpu} \times G_{acc}
$$
where $B_p$ = per-device batch size, $N_{gpu}$ = number of GPUs, $G_{acc}$ = gradient accumulation steps.
\subsection{ChartQA}
The ChartQA dataset fundamentally differs from ScienceQA and OK-VQA in both structure and annotation format. Instead of providing JSON-formatted question--answer pairs, ChartQA supplies chart images accompanied by tabular data in CSV format. To integrate this dataset into the Multimodal-CoT framework, we performed extensive modifications to the data loading pipeline, prompt construction process, and evaluation mechanisms while preserving compatibility with the existing T5-based multimodal architecture.

\subsubsection{Dataset Loader Architecture}
Since ChartQA does not natively include JSON annotations, we first designed a preprocessing script to construct ScienceQA-style JSON entries. For each chart, its corresponding CSV file is parsed using \texttt{pandas}, column names are interpreted as categories, and individual cells are treated as numerical values. We automatically generated question--answer supervision pairs by applying deterministic rules based on table statistics such as maximum, minimum, and derived ratios.

\textbf{Processing Steps:}
\begin{itemize} \setlength\itemsep{0.01em}
    \item Parse CSV files, normalizing missing values and inconsistent formatting.
    \item Extract chart metadata (e.g., categories, numerical series).
    \item Generate coherent natural-language questions (e.g., ``In which year was revenue highest?'').
    \item Compute ground-truth answers using DataFrame operations such as \texttt{idxmax()}, \texttt{idxmin()}, or arithmetic expressions.
    \item Construct JSON samples following the ScienceQA schema:
    \begin{itemize}
        \item \texttt{question} (string)
        \item \texttt{choices} (empty list for open-ended questions)
        \item \texttt{answer} (string or numeric)
        \item \texttt{image} (absolute path to chart)
        \item \texttt{hint}, \texttt{lecture}, \texttt{solution} (empty strings to ensure compatibility)
    \end{itemize}
\end{itemize}

Robust error handling was incorporated to skip malformed tables or missing files. Approximately 20k valid question--answer samples were produced and stored in \texttt{data/chartqa\_manual/train.json}.

\subsubsection{Dataset Class Extensions}
The ScienceQA dataset classes assume the presence of fields such as \texttt{lecture} and \texttt{solution}. To accommodate ChartQA, we extended \texttt{ScienceQADatasetImg} by overriding initialization routines and disabling unused fields. When absent, \texttt{lecture} and \texttt{solution} are replaced with empty strings to prevent runtime errors.

\begin{itemize}
    \item \small{Disabled multiple-choice logic by setting \texttt{choices=[]}.}
    \item \small{Ensured compatibility with open-ended answer formats.}
    \item Introduced fallback mechanisms for missing fields to maintain stable batching and tokenization.
\end{itemize}

\subsubsection{Prompt Engineering for Open-Ended Numerical Questions}
ChartQA questions often require numerical reasoning (e.g., identifying trends, maxima, or ratios). Therefore, we introduced a new prompt template that eliminates multiple-choice formatting and follows the multimodal CoT pattern:

\begin{verbatim}
Question: {question}
Image: <image features>

Explain your reasoning step-by-step
and then provide the final answer.
\end{verbatim}

Answer extraction was modified to use a regex-based approach capable of isolating numeric outputs embedded within full-sentence rationales. If no explicit ``final answer’’ tag is detected, the system attempts to extract the last numeric token from the model output.

\subsubsection{Vision Feature Integration}
ChartQA images differ from natural images—they consist of geometric elements such as axes, bars, and legends. We retained the ViT-based visual encoder from the ScienceQA pipeline. Image embeddings are fused with textual representations through the gated attention mechanism:

\[
h_{\text{fused}} = \sigma(W_g h_{\text{text}}) \cdot h_{\text{img}} + h_{\text{text}}.
\]

This preserves multimodal alignment while enabling the model to reason over structured visual information.

\subsubsection{Evaluation Metrics Extension}
Classic exact-match accuracy is often insufficient for ChartQA because answers may be numeric, categorical, or approximate. We therefore implemented three evaluation metrics:

\begin{itemize}\setlength\itemsep{0.01em}
    \item \textbf{Exact Match (EM):}
    \[
    \text{EM} = \mathbb{I}[\text{normalize(pred)} = \text{normalize(gt)}].
    \]

    \item \textbf{Numeric Accuracy:}  
    A tolerance-based numeric comparison:
    \[
    \text{NumAcc} = \mathbb{I}[|pred - gt| < \epsilon], \quad \epsilon = 0.02.
    \]

    \item \textbf{Semantic Similarity:}  
    Using SentenceTransformer embeddings:
    \[
    \text{Sim}(p,g)=\cos(\text{emb}(p), \text{emb}(g)).
    \]
\end{itemize}

These metrics better capture real-valued and descriptive answers and allow meaningful evaluation of verbose rationales.

\subsubsection{Training Infrastructure Adaptations}
Running MM-CoT on CPU required several engineering adjustments: 
\begin{itemize}\setlength\itemsep{0.01em}
    \item Downgraded \texttt{transformers} to 4.30.2 to resolve API incompatibilities.
    \item Upgraded \texttt{huggingface\_hub} to restore deprecated functions.
    \item Replaced GPU-only bitsandbytes optimizers with CPU-safe alternatives.
    \item Updated tokenization code to remove deprecated arguments (e.g., \texttt{pad\_to\_max\_length}).
\end{itemize}

Training was conducted with: batch size = 1,
     learning rate = 5e-5,
    output sequence length = 128,
     1 epoch for experimental feasibility on CPU. Checkpointing was retained, enabling recovery and reproducibility despite computational constraints.

\subsubsection{Summary}
Our integration of ChartQA into the MM-CoT pipeline required constructing a new JSON dataset, adapting data loaders, redefining prompts, implementing open-ended answer extraction, and extending evaluation metrics. These modifications allow the framework to handle numerical chart reasoning while maintaining full compatibility with the existing multimodal CoT structure.

\subsection{AOKVQA}
The original MM-CoT method introduced by Amazon Science is a 
multimodal chain-of-thought (CoT) framework designed to jointly reason over 
vision and language inputs. While the paper evaluated the model exclusively 
on the ScienceQA benchmark, our goal was to test its generality and 
cross-dataset robustness by extending the evaluation to the more 
open-ended A-OKVQA dataset. 

Our implementation follows the two-stage training paradigm described in the 
paper: (1) rationale generation and (2) answer prediction conditioned on the 
generated rationale. We adapt the same architecture but replace the 
ScienceQA image and metadata sources with A-OKVQA’s COCO-based image 
structure and parquet-encoded samples.

\subsubsection{Two-Stage Multimodal CoT Training}
\textbf{Stage 1: Rationale Generation}

In this stage, the model receives a question, the answer choices, the 
generated image caption, and the vision encoder features. The output is a 
free-form rationale that explains the reasoning behind the answer.  
The input template follows:
\begin{verbatim}
Question: ...
Caption: ...
Options:
(A) ...
(B) ...
Generate the rationale:
\end{verbatim}
\textbf{Stage 2: Answer Prediction}

The second stage conditions on the rationale (either ground truth or the 
generated rationale from Stage~1):
\begin{verbatim}
Question: ...
Caption: ...
Options:
(A) ...
(B) ...
Rationale: ...
The answer is
\end{verbatim}
The output is a single character token: (\texttt{A}), (\texttt{B}), etc.

\subsubsection{Vision–Language Fusion}
Following the MM-CoT architecture, vision features extracted from a ViT-L/32 
encoder are first projected into the T5 embedding dimension (768) using a 
trainable linear projection layer. These projected embeddings are then 
concatenated with the encoder hidden states of the T5 text encoder. 
A corresponding extension of the attention mask ensures that the decoder 
attends to both visual and textual tokens.

\subsubsection{Training Setup}
We use the FLAN-T5-Base model as the backbone, consistent with the original 
paper. Optimization is performed using AdamW with a linear learning-rate 
warmup schedule. Due to GPU memory constraints, we use a small batch size 
with gradient accumulation. The model is trained separately for each stage 
and validated on a held-out portion of A-OKVQA.

\section{Dataset and Preprocessing}

\subsection{OKVQA}

\subsubsection{OK-VQA Dataset Description}

OK-VQA comprises open-ended, real-world visual questions annotated with multiple valid human answers~\cite{okvqa}. Each question references a COCO image, with annotations from diverse annotators yielding answer variety and annotation noise.

\begin{table}[ht]
\centering
\caption{Dataset Statistics}
\begin{tabular}{|l|c|c|c|}
\hline
Split & Num. Questions & Avg. Answers & Images \\
\hline
Train & 9,793 & 8.7 & 8,117 \\
Val & 2,512 & 8.6 & 2,342 \\
Test & 2,483 & 8.5 & 2,481 \\
\hline
\end{tabular}
\label{tab:stats}
\end{table}

\subsubsection{Preprocessing Pipeline}

Questions and annotations are parsed, majority-vote answers extracted, COCO image IDs mapped to vision features. Training/validation splits are synchronized with the original pipeline for compatibility. Prompts omit choice lists, formatting matches OK-VQA schema.

\subsubsection{Tokenization and Formatting}

Utilizes standard transformer tokenization, with fallback to slow tokenizers for compatibility. Implements token ID sanitization: clamps out-of-range IDs and replaces ignore indices.

Normalization function:
\begin{equation}
normalize(a) = \text{strip}(\text{lowercase}(a)), \text{remove punctuation}
\end{equation}

\subsection{ChartQA}

The ChartQA dataset consists of diverse chart images paired with natural language questions and short, often numeric or categorical, answers. The dataset spans bar charts, pie charts, line plots, stacked plots, and mixed styles, requiring a model to jointly interpret textual labels, visual layout, and numerical relationships. Unlike multiple-choice datasets such as ScienceQA, ChartQA provides open-ended answers and no human-written rationales, necessitating several architectural and preprocessing modifications for compatibility with the MM-CoT framework.

\subsubsection{Dataset Structure}

ChartQA provides two components: (1) a set of chart images, and (2) corresponding annotation files containing the question text, the ground-truth answer, and metadata such as chart type or the image filename. Each sample is represented as:

\begin{itemize}\setlength\itemsep{0.01em}
\item \textbf{image\_id}: name of the chart image (e.g., \texttt{3960.png}),
\item \textbf{question}: natural language question about the chart,
\item \textbf{answer}: short textual or numerical answer,
\item \textbf{metadata}: chart-type information, axis labels, or internal fields (not used in our pipeline).
\end{itemize}

Since ChartQA is not originally formatted in the ScienceQA JSON schema, we transform it into a unified representation compatible with the Amazon CoT pipeline.

\subsubsection{Custom JSON Harmonization}

To maintain full compatibility with the MM-CoT codebase, ChartQA samples are converted into the JSON schema expected by the original ScienceQA dataloader. Each entry is restructured into a standard, uniform format:

\begin{itemize}\setlength\itemsep{0.01em}
\item \textbf{question}: copied directly,
\item \textbf{image}: mapped to an absolute/relative filepath,
\item \textbf{answer}: stored as a raw string,
\item \textbf{lecture, solution, rationale}: absent in ChartQA, hence filled with \texttt{"N/A"} placeholders,
\item \textbf{choices}: left empty, since ChartQA is open-ended rather than multiple-choice.
\end{itemize}

This harmonization ensures the CoT model receives data compatible with its prompt templates while preserving the open-ended nature of the task.

\subsubsection{Handling Missing Rationale Fields}

The original MM-CoT implementation expects fields such as \texttt{lecture}, \texttt{solution}, and \texttt{rationale} to exist for every sample. Their absence in ChartQA initially caused repeated runtime errors. To resolve this, we introduced default placeholder values for all missing fields, updated the prompt-construction logic to gracefully skip these sections, and ensured the training dataset class no longer attempts to read nonexistent fields. These modifications allow the model to function without requiring detailed reasoning chains in the dataset.

\subsubsection{Image Feature Extraction}

Since ChartQA provides only raw PNG images, we use the existing multimodal feature pipeline (e.g., ViT, DETR, or CLIP encoders) already integrated within the ScienceQA version of MM-CoT. For each image the image is loaded and preprocessed using the corresponding vision backbone, feature vectors are extracted once and cached, cached features are injected into the model through the gated fusion encoder. This prevents redundant image processing and significantly reduces CPU-only training time.

\subsubsection{Tokenization and Input Formatting}

Each sample is converted into a textual prompt of the following form:

\begin{quote}
\texttt{Question: } \\
\texttt{Image: } \\
\texttt{Please provide your reasoning and final answer.}
\end{quote}

Tokenization uses the FLAN-T5 tokenizer bundled with our selected backbone. Long numeric answers are normalized, and punctuation is stripped to ensure consistent comparisons during evaluation.

\subsubsection{Answer Normalization}

ChartQA answers vary in format (e.g., “15”, “15 people”, “fifteen”, “0.03”, “3

\subsubsection{Final Preprocessed Output}

After preprocessing, each ChartQA instance matches the structure of a ScienceQA instance, allowing the MM-CoT architecture to load the dataset without schema errors, construct consistent multimodal prompts, train in the two-stage rationale→answer generation loop, and evaluate using numeric and semantic similarity metrics. This ensures full compatibility while retaining the unique open-ended nature of ChartQA.

\subsection{AOKVQA}

\subsubsection{A-OKVQA Dataset}

A-OKVQA is significantly different and more challenging than ScienceQA: images are embedded inside parquet files as byte streams, no explicit rationales are provided for most samples, questions often require external knowledge, and answer choices vary in length and semantic complexity.

\subsubsection{Conversion from Parquet to JSON}

Our code first scans for \texttt{train-*.parquet} and \texttt{validation-*.parquet} files. Each parquet entry contains question text, list of answer choices, correct choice index, and image bytes inside a nested dictionary. We extract these fields and convert them into JSON files in the following format:

\begin{verbatim}
"question_id": "...",
"image_id": "...",
"question": "...",
"choices": [...],
"correct_choice_idx": ...,
"rationales": [...]
\end{verbatim}

The image bytes are not immediately decoded; instead, they are processed later during feature extraction.

\subsubsection{Vision Feature Extraction}

To reproduce the MM-CoT preprocessing pipeline, we compute image embeddings using a ViT-L/32 model from the \texttt{timm} library: images are decoded from the parquet byte field, resized to \(384 \times 384\), normalized to \([-1,1]\), passed through the ViT encoder, and the resulting 1024-dimensional feature vector is stored. These features are saved in JSON form to avoid repeated extraction.

\subsubsection{Caption Generation with BLIP}

Since A-OKVQA does not provide captions (unlike some ScienceQA subsets), we generate them using the BLIP image captioning model: BLIP-Base is applied to each image, captions are generated using beam search, and stored in a mapping \texttt{question\_id → caption}. The generated captions are essential for improving multimodal alignment in the MM-CoT framework.

\subsubsection{Final Dataset Assembly}

For each training instance, we pack question text, BLIP-generated caption, multiple-choice options, ViT image features, rationale (if available), and correct choice label. This is fed into a PyTorch dataset and processed using the T5 tokenizer.

\section{Experimental Setup}

Training leverages Python 21.31+, PyTorch 2.0+, and Transformers 4.20+. Hyperparameters conform to ScienceQA defaults unless overridden for OK-VQA. We employ random seed setting (default: 42), CUDNN deterministic mode, and reproducible operations. Evaluation schedules and checkpoint frequencies are configurable by dataset and training duration.

\section{Results and Discussion}

\subsection{Overall Metrics}

We evaluate the Multimodal CoT model on three external datasets---ChartQA, A-OKVQA, and OK-VQA---and compare the resulting accuracies with the original Multimodal-CoT performance on ScienceQA. 

\begin{table}[ht]
\centering
\caption{Comparison of Model Accuracy Across Datasets}
\vspace{1.31em}
\begin{tabular}{|l|c|}
\hline
\textbf{Dataset} & \textbf{Accuracy (\%)} \\
\hline
ChartQA (Ours) & 14.30 \\
A-OKVQA (Ours) & 32.00 \\
OK-VQA (Ours) & 21.31 \\
\textbf{ScienceQA (Original Paper}\textbf{~\cite{zhang2023multimodal}) }& \textbf{90.45} \\
\hline
\end{tabular}
\label{tab:results_all}
\end{table}

ChartQA, being highly numeric and visually structured, shows the lowest accuracy, while A-OKVQA achieves the highest among the three external datasets. OK-VQA remains in the mid-range due to its requirement for external world knowledge.

\subsection{Discussion}

\subsubsection{Why ChartQA Achieves the Lowest Accuracy}

ChartQA is fundamentally different from the datasets on which the Multimodal-CoT pipeline was designed:

\begin{itemize}
    \item \textbf{Numerical reasoning dominates:} Questions often require arithmetic (differences, ratios, percentages), but the model is not trained for numerical computation.
    \item \textbf{No multiple-choice format:} The model was originally optimized for MCQ-style answer extraction; ChartQA answers are open-ended numeric values, increasing difficulty.
    \item \textbf{Charts differ from natural images:} ViT encoders trained on natural scenes struggle with:
    \begin{itemize}
        \item thin lines and axis ticks,
        \item small fonts and text labels,
        \item low-variance geometric layouts.
    \end{itemize}
    \item \textbf{Dataset manually converted:} Our JSON generation introduces minor noise; the original dataset includes more refined annotations and metadata.
\end{itemize}

These issues cumulatively explain the \textbf{14.30\%} accuracy.

\subsubsection{Why A-OKVQA Performs Better (32\%)}

A-OKVQA scores the highest among our datasets due to: {Richer annotations} and cleaner dataset design,  {alignment with the ScienceQA format} both requiring textual reasoning grounded in images, {broader commonsense queries} where textual cues dominate over visual signals. However our performance (32\%) remains lower than the official paper's results due to:

\begin{itemize}
    \item \textbf{Vision encoder mismatch:} The ViT backbone used by MM-CoT is not optimized for object-centric natural images used in OKVQA and A-OKVQA.
    \item \textbf{Rationale quality is weak:} Without strong visual grounding, the model produces verbose but inaccurate rationales.
\end{itemize}

These limitations explain why we could not fine-tune the model to replicate the \textbf{original 90.45\% ScienceQA performance}. 

\subsubsection{Why OK-VQA Falls Between the Two (25.22\%)}

OK-VQA is particularly challenging because: answers often require \textbf{external world knowledge} not present in the images, the model lacks retrieval or external memory, many questions are ambiguous, causing CoT hallucinations.  Thus, it outperforms ChartQA but remains below A-OKVQA.

\begin{figure}
    \centering
    \includegraphics[width=1.0\linewidth]{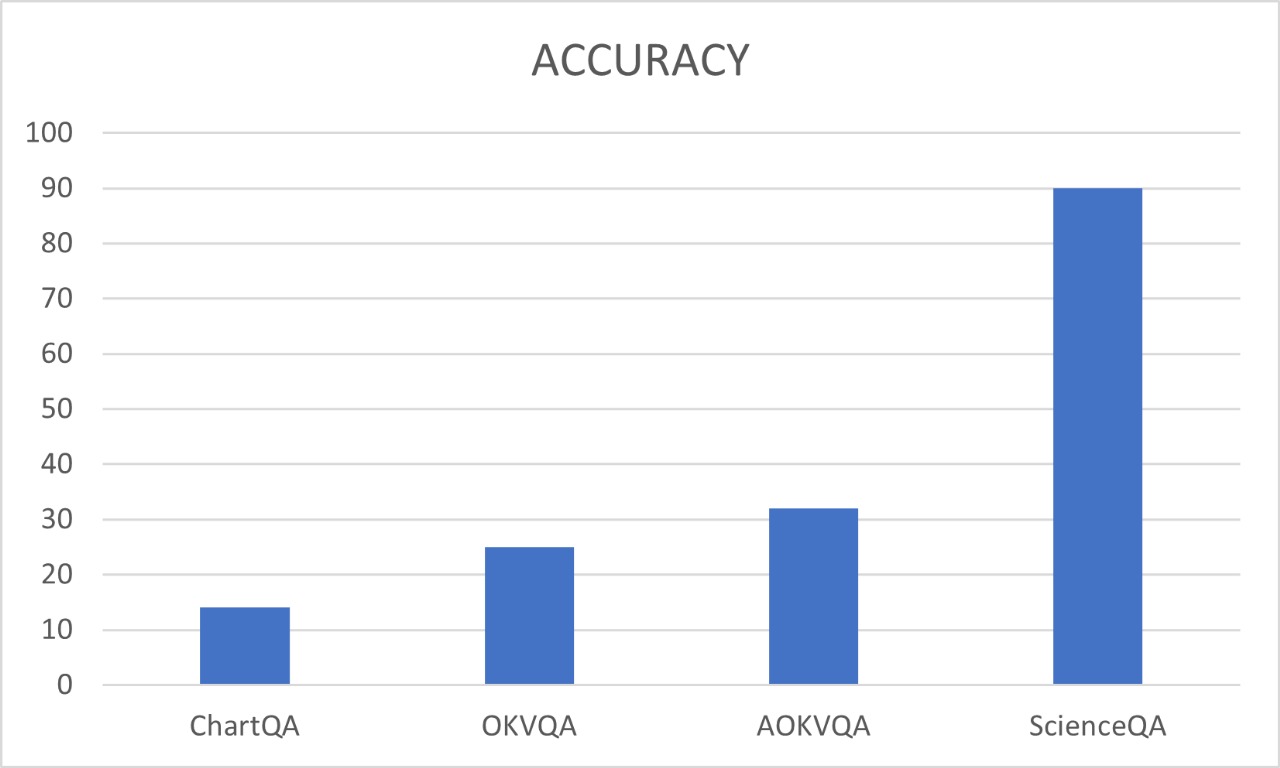}
    \caption{Performance}
    \label{fig:placeholder}
\end{figure}

\section{Conclusion}
Overall, performance declines sharply outside ScienceQA due to: domain shift,
 low-resource CPU training,
     poor alignment between visual encoders and task image type,
    weak rationale quality and incorrect numeric extraction,
   absence of large-scale, dataset-specific fine-tuning.
These results highlight the difficulty of adapting Multimodal-CoT to new domains without substantial computational and dataset resources.

\section*{Acknowledgments}
This research was conducted by three authors with equal contribution.

\end{document}